# Towards LLM-based Autograding for Short Textual Answers

Johannes Schneider[1], Bernd Schenk[1] and Christina Niklaus[2]
[1]Department of Computer Science and Information Systems, University of Liechtenstein, Vaduz, Liechtenstein
[2]School of Computer Science, University of St.Gallen, St.Gallen, Switzerland
{Johannes.Schneider,Bernd.Schenk}@uni.li, Christina.Niklaus@unisg.ch

Abstract: Grading exams is an important, labor-intensive, subjective, repetitive, and frequently challenging task. The feasibility of autograding textual responses has greatly increased thanks to the availability of large language models (LLMs) such as ChatGPT and because of the substantial influx of data brought about by digitalization. However, entrusting AI models with decision-making roles raises ethical considerations, mainly stemming from potential biases and issues related to generating false information. Thus, in this manuscript we provide an evaluation of a large language model for the purpose of autograding, while also highlighting how LLMs can support educators in validating their grading procedures. Our evaluation is targeted towards automatic short textual answers grading (ASAG), spanning various languages and examinations from two distinct courses. Our findings suggest that while "out-of-the-box" LLMs provide a valuable tool to provide a complementary perspective, their readiness for independent automated grading remains a work in progress, necessitating human oversight.

## 1 INTRODUCTION

Large language models like ChatGPT are said to be "foundational" for many tasks having led to a widespread impact across both industry and academia (Schneider, Meske, et al., 2024). However, artificial intelligence, including LLMs, is hard to understand (Longo et al., 2023; Meske et al., 2022) and suffers from security concerns (Schneider & Apruzzese, 2023). Experts also perceive substantial risks associated with LLMs and have advocated for a development moratorium on such technologies (Future of Life, 2023). In academia, LLMs are used as a tool by researchers and students to such an extent that researchers themselves have called on journals to clarify the allowable extent of AI-generated content in scholarly papers (Tang, 2023), leading to the publication of guidelines for incorporating AI in the paper-writing process(Aczel & Wagenmakers, 2023). Ethical concerns have also been raised for education (Yan et al., 2023) and children (Schneider, Kruse, et al., 2024). LLMs like ChatGPT have been commonly compared against students in various disciplines – especially with respect to their capability to pass exams. While some reports have indicated inferior performance than a master's graduate in mathematics (Frieder et al., 2023), other instances showcase a successful completion of an introductory physics course (Kortemeyer, 2023), as well as the passing of numerous law school exams (Choi et al., 2023). However, it is important to acknowledge the existence of limitations in the LLMs. These models can exhibit biases, discrimination, and factual inaccuracies (Borji, 2023). Consequently, there arises doubt regarding their suitability in education. In particular, the necessity for human verification has been emphasized as a pressing research priority (van Dis et al., 2023) and the topic of human agency is also debated on a regulatory level (EU, 2020). Especially, high stakes decisions require careful analysis before AI can be utilized. Grading of exams is a high-stakes situation, as errors in grading can cause students to fail an entire class, possibly causing a year-long delay in their education, separation of peers, etc. This, in turn, can lead to both financial and psychological strain.

As such it seems natural and even necessary to assess the suitability of LLMs for supporting exam grading and to reflect upon adequate ways to include them in the grading process while mitigating their risks. To this end, we seek to contribute to two intertwined research questions: (a) How can LLMs support educators in the grading process of exams? (b) What are issues and concerns when using LLMs to support grading?
Our focus is on Automatic Short Answer Grading (ASAG), i.e., student replies are (short) textual

answers (i.e., one or a few paragraphs). We use an LLM, i.e., ChatGPT to assess the instructor's answer, a student's answer in general as well as a student's answer with respect to the instructor's answer as illustrated in Figure 1. In our experimental evaluation, we used two exams from two educators.

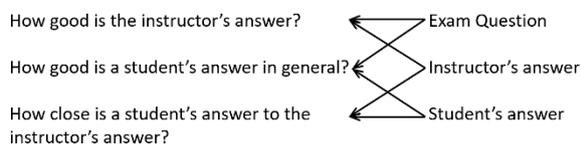

Figure 1: The three questions an LLM answers to support the grading(left) and the information used by the LLM (right) to answer them.

While we implicitly assess the possibility of automatic grading, our target is (i) to improve the grading process rather than simply automating it and (ii) to uncover shortcomings (and possible mitigations) of LLMs for this purpose. We seek to employ LLMs as a second opinion that might pinpoint obvious flaws in the grading, i.e., due to sloppiness, in the grading as well as provide a more general view on possible answers in order to avoid bias like accepting only correct answers that have been discussed in the course.

## 2 METHODOLOGY

We use an LLM, i.e., ChatGPT (GPT 3.5, June/July 2023 versions), to assess (i) answers by the educator, (ii) answers of students to exam questions in general, and (iii) answers of students compared to the instructor's answer (see Figure 1).

That is, to assess an instructor's answer, an instructor must be able to define an answer for each exam question that constitutes the optimal response from her/his perspective. We shall elaborate on lifting this requirement and allow for multiple possible answers per question in the discussion section.

Our assessment is both qualitative and quantitative. That is, we highlight a few prompts that surfaced surprising issues (such as lack of robustness, i.e., sensitivity to minor variations in prompts), but we also quantify how much the LLM deviates from the educator across all graded answers. To this end, we ask the LLM to categorize its assessment, i.e., each LLM response should contain a category such as "Good", "Ok.", or "Bad" and an explanation of the chosen category. In turn, we also categorize the educator's responses. This allows us to compare the categorization of the LLM and the educator, e.g., to assess if both rate a student answer as "good", which we elaborate in the last subsection. We experimented with a few different prompts but we report only the ones used in the end.

### 2.1 Assessing the educator's answer

An educator should have a good sense of different answers to exam questions. However, it is often difficult to anticipate all arguments and answers. For example, an educator might be primed to answers that are based on the material taught in class but students might include elements with different but nonetheless valid viewpoints. We utilize the LLM to provide a more general view on the question without explicit knowledge on the lecture material. We also ask the LLM for feedback on the developed educator's solution. The prompt has the following structure for a question Q and the educator's answer A.

*Here is a question: Q . Here is an answer: A. How good is the answer to the question? Start the reply with one of the following: Extremely good., Very good., Good., Ok., Bad., Very bad., Extremely bad. Explain the choice. Explain also what is missing.*

Example Prompt (1) to LLM: *Here is a question: What is the difference between single linkage and average linkage (hierarchical) clustering? . Here is an answer: The two differ in distance metric used to cluster. Single linkage: Merge two clusters based on minimum distance between any two points; Tendency to form long chains; Average linkage: merge two clusters based on average distance between any two points; tendency to "ball" like clusters;. How good is the answer to the question? Start the reply with one of the following: Extremely good., Very good., Good., Ok., Bad., Very bad., Extremely bad. Explain the choice. Explain also what is missing.*

Response of LLM: *Good. The answer provides a clear and concise explanation of the difference between single linkage and average linkage clustering. It accurately states the different distance metrics used and describes the tendencies of each clustering method. One potential improvement is to provide a more detailed explanation of the advantages and disadvantages of each method. Additionally, it would be helpful to include an example or illustration to further clarify the concepts.*

### 2.2 Assessing the student's answer

We let the LLM assess the student's answer to a question. The prompt has the following structure for a question Q and a student answer A.

*Here is a question: Q . Here is an answer: A. How good is the answer to the question? Start the reply with one of the following: Extremely good., Very*

*good., Good., Ok., Bad., Very bad., Extremely bad. Explain the choice.*

Example Prompt to LLM: *Here is a question: What is the difference between single linkage and average linkage (hierarchical) clustering? . Here is an answer: In single linkage, we compare the two closest data points (the ones with minimal distance) from two separate clusters. In average linkage, we compare all the data points from a cluster with all the datapoints from another cluster and get an average distance. .How good is the answer to the question? Start the reply with one of the following: Extremely good., Very good., Good., Ok., Bad., Very bad., Extremely bad. Explain the choice.*

Response of LLM: *Good. The answer provides a clear and concise explanation of the difference between single linkage and average linkage clustering. It accurately describes how single linkage compares the closest data points from two separate clusters, while average linkage compares all data points from a cluster with all data points from another cluster and calculates the average distance. The answer effectively addresses the question and provides a solid understanding of the topic.*

## 2.3 Comparing student's and educator's answer

We compare the student's answer with the educator's answer. The LLM assesses the similarity between the student answer and the educator's answer, i.e., the optimal answer from the perspective of the educator. The LLM only obtains the educator's and the student's answer as input but not the question (see also Figure 1). We found that when the question is added, the assessment of ChatGPT sometimes tended to neglect the optimal answer (although the prompt clearly stated that the task was to compare the student reply and the educator's answer).

Thus, the LLM provides a second opinion like another grader that had also knowledge on the course in the form of what is the best answer given the taught material, i.e., the educator's answer. Thus, in contrast to the prior assessments that were seeking for a more general perspective, here, we ultimately seek to assess, if an educator's grading matches that of the LLM given knowledge of the best response. That is if both should stick to a pre-defined "ground truth" in the form of the educator's answer. We shall also discuss how to relate the grading of the LLM and the educator quantitatively in the next section. As a prerequisite to derive quantitative statements, we ask for a category but also for an explanation to assess qualitatively. More concretely, the prompt has the following structure for a student answer A and the (assumed to be optimal) educator's answer E.

*Here is an answer: A . Here is the optimal answer: E. How close is the answer to the optimal answer? Start the reply with one of the following: Very close., Close., Somewhat close., Somewhat distant., Distant., Very distant.. Explain the choice.*

Example Prompt: *Here is an answer: In single linkage, we compare the two closest datapoints (the ones with minimal distance) from two separate clusters. In average linkage, we compare all the datapoints from a cluster with all the datapoints from another cluster and get an average distance. .*

*Here is the optimal answer: The two differ in distance metric used to cluster. Single linkage: Merge two clusters based on minimum distance between any two points; Tendency to form long chains; Average linkage: merge two clusters based on average distance between any two points; tendency to "ball" like clusters; How good is the answer to the question? Start the reply with one of the following: Extremely good., Very good., Good., Ok., Bad., Very bad., Extremely bad. Explain the choice. Explain also what is missing. What is the difference between single linkage and average linkage (hierarchical) clustering?*

*How close is the answer to the optimal answer? Start the reply with one of the following: Very close., Close., Somewhat close., Somewhat distant., Distant., Very distant.. Explain the choice.*

Response of LLM: *Very close. The given answer effectively highlights the key differences between single linkage and average linkage in hierarchical clustering. It mentions that single linkage does not allow linkage between different groups, while average linkage splits the dataset into average-sized groups. Overall, the given answer accurately addresses the main distinction, making it very close to the optimal answer.*

## 2.4 Highlighting deviations of LLM and educator

The prior assessments are helpful, but require a lot of effort for the educator, i.e., the educator must manually assess all responses of the LLM. It might be desirable to focus more on large differences between human grading and the reply of the LLM. Furthermore, quantifying differences allows for a quantitative assessment, which makes it more explicit how well aligned the educator's grading and the LLM's grading are. We assume that an educator has provided a score in some range [0,*maxPts*], where *maxPts* gives the maximum points attainable by a student. For easier interpretability, we scale the scores to be between 0 and 1. We denote the scaled points given by the human, i.e., educator, as $p_h$. We also compute a score given by the LLM denoted as $p_L$ based on the LLM's responses. We compute correlations to understand the alignment between human and LLM. We sort the outcomes based on how different the LLM's ratings are from the educator's grading, which allows an educator to focus on the largest differences. The deviations are computed as follows. We assume that if the LLM replies for a given student answer and the educator's answer with "very close" then that student should have gotten maximal points, whereas if the response is "very distant" the student should have gotten no points. Similarly, if the LLM rates a student answer for a given exam question as "extremely good" then the student should have gotten maximal points. If it is rated "extremely bad" then no points are given. Inbetween, we scale accordingly, i.e., we linearly interpolate to obtain the points an LLM would assign for other categories. We also scale by *maxPts* to have a normalized value between 0 and 1 for the LLM. We denote the scaled points by $p_L$. We then sort outcomes based on the gap between the human's and LLM's scores $| p_h - p_L |$. This allows to focus on responses where the educator and LLM disagree the most, i.e., we sort the LLM responses from largest to lowest differences. For easier comparison we show both the educator's (or student's) response and the LLM's response. This enables us to read the answer of the human, i.e., student or educator, and then the reply of the LLM. If the two responses disagree significantly, and indeed the LLM had a valid point, then an educator should alter his/her grading.

An example output is:

*Gap: 0.9  LLM Pts p_l: 0.1  Human Pts p_h: 1.0*

*Answer Human: When using single linkage in hierarchical clustering, the clusters are made with the shortest distance between the data point the closest cluster. Average linkage on the other hand takes the maximum distance and the minium distance to each cluster, evaluates the average and then builds the clusters upon these average distances.*

*Answer LLM: Somewhat close. The answer provides a basic understanding of the differences between single linkage and average linkage in hierarchical clustering. It correctly states that single linkage considers the shortest distance between a data point and the closest cluster, whereas average linkage calculates the average distance to each cluster. However, it does not fully explain how average linkage clusters are formed based on these average distances.*

## 3 ASSESSMENT SETUP

We assess two exams administered by two different educators. One exam was in English at the master level in data science with 21 participants. It consisted of 16 questions and had a total of 90 points. It lasted 90 minutes. The other was in German at the bachelor level in an information systems introductory course with 34 participants. There was a multiple choice part that was not considered, i.e., we were left with 3 questions each worth 10 points resembling 10 minutes of processing time per question, i.e., the replies were expected to be longer than for the questions of the data science exam. The types of questions covered all categories according to the revised Bloom's taxonomy of educational objectives(Anderson & Krathwohl, 2001). The taxonomy ranges from simple, concrete to complex, abstract questions. Our exams contained some simple concrete questions related to understanding and remembering, such as providing definitions and examples. More complex, abstract questions consisted, for example, of evaluating different approaches for a practical problem. We read through all of ChatGPT's responses.

## 4 FINDINGS

We first discuss overarching findings before elaborating on each of the three questions in Figure 1.

**The LLM replies generically.** ChatGPT tends to assess in a mechanistic generic manner rather than looking at content. It might respond like "There is not sufficient detail" rather than pointing to specific details that are missing.

**The LLM and human assessments differ strongly.** The correlation between human and LLM's judgments is small. Generally, the LLM's judgments have a strong tendency to the middle, e.g., most are "ok" or "good" despite strong variation in the quality of student replies.

**The LLM can help to make sense of hard to understand answers.** The LLM provided a more open and less negative view on responses suffering from poor language. Thus, the assessment was particularly positive for students with poor (English) language skills, as ChatGPT tended to rate them comparatively better to an educator. That is, a human might rate them poorly because the answer is difficult to understand or remains unclear due to grammatical ambiguities or poor wording. We also found that it was sometimes easier to make sense of a student reply after reading ChatGPT's assessment. Furthermore, commonly specific concepts tied to a set of keywords are accepted or looked for. If students do not provide any of these but rather a lengthy and verbose reply, there is a higher risk that possibly correct though convoluted arguments are overlooked. We found that ChatGPT's assessment can be helpful, since it can transform hard to grasp answers into a more concise phrasing and its responses follow an expected structure, which is fast to process for an educator.

***The LLM can drastically change its assessment due to minor changes in answers***. Additional content that is strikingly wrong though not related to the question (or answer) can lead to dramatic changes in judgements by the LLM. For illustration, we appended to the answer of the student used in the example prompt (1) either of the following three options:

(i) *3\*5=7*,

(ii) *the cat sits on the mattress*,

(iii) *3\*5=7, the cat sits on the mattress*;

ChatGPT judged two of them equivalently as the original prompt (1), i.e. as good. For (ii) and (iii) it would mention that the answers contain irrelevant information, but (ii) is still judged as good by the LLM, while the LLM judged the response (iii) as "very bad".

***The LLM favors vague content and fails to recognize contradictions.*** Generally, replies with vaguely related content, which might be deemed irrelevant or even incorrect by a human grader, is rated more favorably by the LLM than by human graders. We also found that ChatGPT can fail to distinguish contradicting statements. We appended to prompt (1) either of the following:

(i) *Complete linkage uses the minimum distance between any two points in clusters. Density based clustering relies on computing the number of points, possibly organizing them in a search tree or a list.*

(ii) *Complete linkage uses the maximum distance between any two points in clusters. Density based clustering relies on computing point densities, e.g. points for a fixed volume, for the volume for a fixed set of points.*

Note, the words minimum and maximum are switched in (i) and (ii). The LLM judged (i) and (ii) equally, although they obviously contain contradicting statements and information not being asked for.

***The LLM misunderstands questions.*** ChatGPT can suggest to provide information that can obviously be ruled out as being asked for. For the question "What are advantages of a decision tree?" (and a student's answer) the LLM's reply included "However, what is missing from the answer is a mention of some potential disadvantages or limitations of decision trees."

***The LLM's grading criteria are language sensitive.*** We applied the same prompt patterns for both exams, i.e., we utilized the English prompt pattern for the German exam. While at first, this did not seem to pose a problem, we found that ChatGPT occasionally provides a lower rating giving as reason that (German) texts contain grammar and spelling issues, but this would not happen for answers in English.

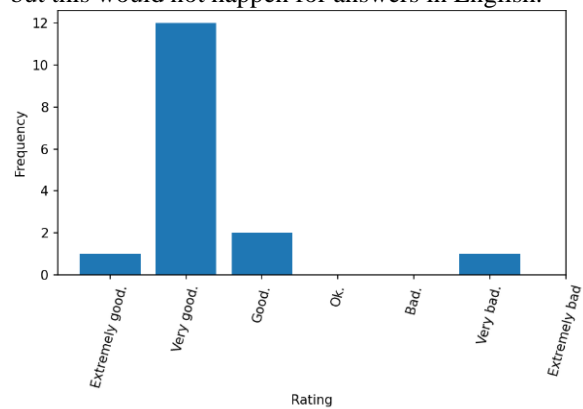

Figure 1: Ratings of educator's answers by LLM

## Findings on assessing the educator's answer by the LLM

We read through all of ChatGPT's responses. None of them led to any changes of the human-crafted responses. Most suggested improvements were generic, e.g., related to giving more details, an example and sometimes visualization or limitations. ChatGPT's responses were quite sensitive to the phrasing (and potentially other factors). For example, omitting the term "Explain also what is missing." changed the LLM's response for one reply from "very bad" (see Figure 2) to "good", while still giving mostly the same reasoning. Overall, Figure 2 suggests that the educator provided mostly "very good" answers and no answer was below "good" (at least when slightly changing the prompt as mentioned before).

## 4.1 Findings on assessing the student's answer in general and relative to the educator's answer

Here, the LLM had more impact on the grading. That is, we made minor adjustments after reading through the LLM's assessment. The adjustments were made due to two types of replies: First, the LLM would rate student answers (more) positively that were not part of the lecture material and also not directly being asked for. For example, we asked "Which two of the factors 'data, compute and algorithms' are most important for the rise of GPT-3 (around 2020) and ChatGPT in 2022 and other generative AI models?" Some students responded that one factor was media coverage and accessibility due to its public release. ChatGPT rated the responses of these students positively, although (1) the question explicitly restricts the factors to be discussed and (ii) the importance of media coverage is debatable – at least for GPT-3. That is, in the lecture, it was mentioned for ChatGPT that its public release led to widespread adoption and a surge in media coverage, but not so much for GPT-3. GPT-3 was covered less in the media, and it was less accessible, i.e., only (some) scientists got access.

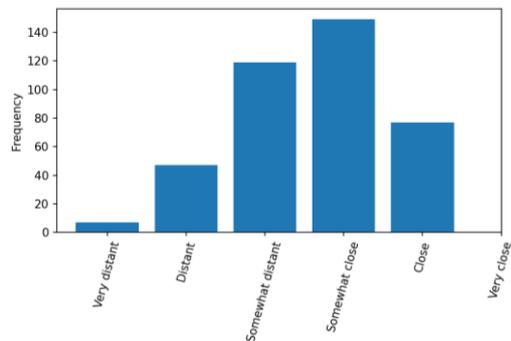

Figure 2: Ratings of students' answers by LLM

Second, the LLM would more positively rate replies with poor English. That is, the LLM's interpretation of the answer made the student answer more understandable. For an educator, the quality of the answer needs to exceed a minimum level of proficiency to be understood. Comprehensibility is generally a factor influencing grading. Educators are not supposed to make too many assumptions about what a student wants to say (i.e. interpret) but they have to stick with the incomprehensible answer and grade accordingly.

Overall, we found that any judgement of the grading after consulting the LLM was preceded by considerable reflection and debates, and it was not evident whether the differences of the LLM should really be considered.

Interestingly, using an answer set of an exam conducted in German, the LLM incorporated errors in spelling and grammar in the feedback and downgraded answers of poor language quality.

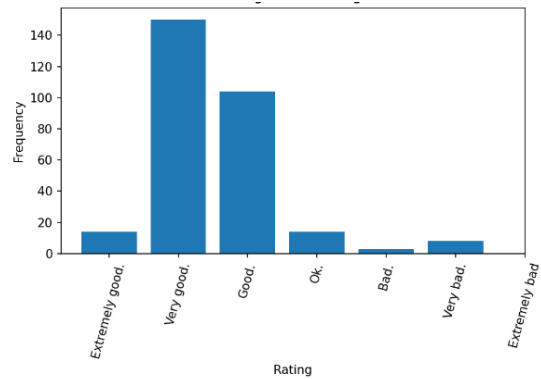

Figure 3: Comparison of students and educator's answers by LLM.

The LLM tended to rate most student responses as "good" or "very good" (Figure 2), i.e., there was little differentiation. This is in stark contrast to the rating of the educator (Figure 3). The educator scored many answers with maximum or minimum points but he/she also assigned commonly points in-between the two extremes. The extremes were mostly common for short and easy answers with few points.

When it comes to assessing similarity between the educator's and the students' answers the LLM gave somewhat more diverse replies. However, overall alignment was poor. The Pearson correlation between the LLM's similarity assessment $p_L$ and educator's grading $p_h$ was close to 0.

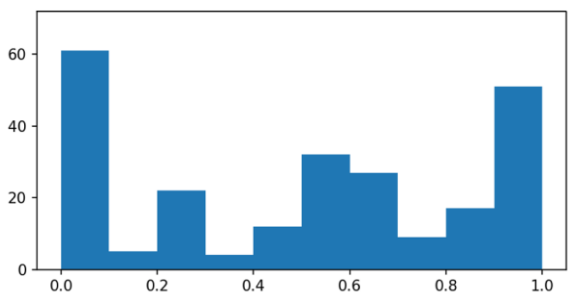

Figure 4: Distribution of frequency (y-axis) of normalized points by educator (x-axis)

# 4 DISCUSSION

We set out to assess LLMs for autograding, primarily as a second opinion as for high stakes decision regulation also demands human oversight. Specifics of the course have not been provided to the LLM, e.g., the course material for which the exam was made for. That is, the LLM lacked any lecture specific context. It relied on world knowledge incorporated in its training data. Thus, discrepancies between the judgments of the LLM and lecturer are expected, e.g., as many terms are defined differently in other contexts and fields. Such contextualization of terms seems to be an essential part of teaching and learning and allows students to establish different perspectives on issues. However, for grading, we believe that the lack of context by the LLM can be highly valuable, as it provides a strongly complementary view that aims to avoid strong biases of a lecturer towards the lecture material. Still, this also hints that grading (or possibly even exam questions) derived by an LLM given access to the course material could provide a welcome addition to be investigated in future work.

We also faced a number of practical issues when using LLMs. For example, the LLM's replies would not always follow the given structure, i.e., ChatGPT would reply with any of the asked for words "Very good", "Good" etc. but started the reply with some other sentence. This problem can often be mitigated by providing a few examples of inputs and desired outputs (in-context learning). However, doing so means additional work for the educator, increases response time of the LLM and also costs, i.e., longer input prompts imply higher costs for commercial LLMs.

Our experimentation with prompting surfaced trade-offs. For example, when comparing the student's and the educator's answer, we tested prompts that included the question (as well as the answers of the student and educator) and prompts that did not. We found that without the question, ChatGPT's assessment sometimes tended to include aspects that were rather unrelated to the question. If the question was added, the assessment of ChatGPT sometimes did not consider the educator's answer.

It is also tempting to use LLMs for fully automatic grading. However, from our experience this should not be undertaken at the current point in time since there is very strong disagreement between gradings of educators and LLMs. That is, they perform significantly worse in judging questions than in providing responses. This might be improved using techniques such as in-context learning, i.e., providing examples on how answers should be graded, or fine-tuning LLMs specifically towards autograding. However, first experimentation did not yield the hoped performance boost. In general, finding the best prompts for grading is non-trivial and responses could be sensitive to the slightest changes in phrasing. Grading should be robust, fair, and consistent. Accordingly, the achievement of competency levels of students should be assessed as independently as possible of individual course delivery, of lecturers and examiners, and of the performance of other students in an exam. ChatGPT did not (yet) meet these requirements in our evaluation.

We assessed the idea to focus on answers where the LLM and the human showed largest discrepancies. However, unfortunately, ChatGPT's rating was not too well-aligned with that of the educator. Furthermore, if not all answers are checked (but only those with large differences), biases in the LLM might further impact the grading by leading to a bias in which answers are looked at (again) by a human grader. Furthermore, biases also appear as "misleading clues". If, for example, the LLM judges arguments identical to a human, except for an argument A, then students using A are more likely to show a large gap (even if aligned with the educator) and thus being assessed by the educator.

One assessment within our work assumed that an educator provides a single answer to a question. In principle, a question might permit fairly different valid answers. However, it is not hard to allow for multiple responses, i.e., an educator could define various answers that might even be contradictory. We could then compare a student's answer with all of the educator's responses and focus on the educator's response that is deemed closest by the LLM. However, specifying answers becomes more difficult the more open-ended a question is, i.e., the more knowledge should be applied and transferred, as opposed to simply replicating knowledge.

From an ethical point of view, one might also debate whether changes due to LLMs should only improve grades. That is, LLMs should not be allowed to fail a student, as punishing innocent people can be seen as worse than rewarding people not having deserved it. Furthermore, using an LLM as "a second opinion" might also provide a false sense of security.

## 5 FUTURE WORK

We might add a more explicit grading scheme that aims to identify specific aspects in the answer, i.e., "Is this concept in the answer?" (If not deduct x points). Furthermore, a fine-tuned LLM towards grading might lead to better outcomes than relaying on prompting. To this end, large number of graded exams would be needed. While graded exams already exist, sharing them is non-trivial as responses might have to be anonymized to comply with privacy regulations.

LLMs might also be useful for exam development, i.e., assessing questions prior to posting an exam. One might also provide access to the lecture material to the LLM to assess gradings. This might uncover more minor issues in the grading scheme, but might not help so much in uncovering general issues. In this study, we used LLM only in grading answers on questions that have been formulated by lecturers. It would be interesting to test the end-to-end support by LLMs in designing a lecture, including the selection of topic areas, creating the lecture material, and preparing and assessing the exam.

## 6 RELATED WORK

The manual grading process involves a labor-intensive evaluation of students' responses, requiring expertise and careful judgment to assign appropriate scores. Thus, to assist educators in reducing the time and effort spent on grading, there is a growing interest in leveraging AI-driven correction aids (Basu et al., 2013; Condor et al., 2021). When comparing the conventional teacher's judgement ("human scoring") to the capabilities auf automatic feedback and assessment tools ("machine scoring"), we encounter distinct strengths along various quality criteria (Seufert et al., 2022), i.e., AI can support objectivity, reliability, validity and comparative values and standards.

The evolution of assessment methodologies is currently exploring hybrid solutions that harness the strengths of both mechanisms. These developments, such as AI-based assistants for assessment and learner feedback, hold promise for the future education, offering more efficient and objective evaluation processes while maintaining the depth of understanding provided by human judgement (Saha et al., 2018; Schneider et al., 2023). A few works have also assessed the value of feedback through autograding, e.g., (Vittorini et al., 2020) also assesses the value of feedback provided by the autograder for students. (Li et al., 2023) investigated the effects of AI grading mistakes on learning showing that, in particular, marking wrong answers as right had a negative impact on learning

We concentrate on the field of Automatic Short Answer Grading (ASAG) (Burrows et al., 2015). It deals with grading student answers, typically ranging from a phrase to a paragraph. ASAG also covers the grading of open-ended answers (Baral et al., 2021). The primary focus in ASAG is on content quality rather than the writing style and structure emphasized as in automatic essay scoring (AES) (Dikli, 2010).

For ASAG, prior work has mostly relied on BERT as a large language model (Baral et al., 2021; Haller et al., 2022; Schneider et al., 2023; Sung et al., 2019; Zhang et al., 2022). Schneider et al. (2023) investigated how LLMs such as BERT suffer from trust issues that might be partially mitigated by only automatically grading answers, if the LLM is certain about its grading.

While LLMs can provide justifications for their decisions without any additional work, dedicated methods for enhancing explainability have been evaluated in the realm of automatic grading systems (Kumar & Boulanger, 2020). Efforts have also been made to tackle the limitations of AI in terms of fairness in the context of automatic grading (Madnani et al., 2017) and, more generally, ethical issues related to LLMs in education (Yan et al., 2023).

## 7 CONCLUSIONS

The integration of LLMs into academic settings have become an undeniable reality. These models possess remarkable linguistic capabilities, coupled with unexpected reasoning abilities. Yet, using LLMs such as ChatGPT "out-of-the-box" to support grading requires great care due to a multitude of issues such as sensitivity to minor changes in answers and lack of concise reasoning, which is also reflected in poor alignment with human graders. Despite these limitations, LLMs currently offer a valuable resource that provides a supplementary viewpoint with minimal effort.